\pdfoutput=1

\documentclass[11pt]{article}

\usepackage{naacl2021}

\usepackage{times}
\usepackage{latexsym}

\usepackage[T1]{fontenc}

\usepackage[utf8]{inputenc}

\usepackage{url}
\usepackage{hyperref}
\usepackage{caption}
\usepackage{graphicx}
\usepackage{booktabs}
\usepackage{subcaption}
\usepackage{mathtools, cuted}
\usepackage{dsfont}
\usepackage{array}
\usepackage{enumitem}
\usepackage{amsmath,amsfonts,amssymb,amsthm, bm}

\DeclareMathOperator*{\argmax}{arg\,max}
 
\usepackage{microtype}

\usepackage{fancyhdr}
\title{Zero-Shot Generalization using \\ Intrinsically Motivated Compositional Emergent Protocols}

\author{
  Rishi Hazra\thanks{\;\;Equal Contribution}, \,\, Sonu Dixit\footnotemark[1],  \,\, Sayambhu Sen \\
  \{\texttt{rishihazra, sonudixit, sayambhusen}\}\texttt{@iisc.ac.in} \\
  Indian Institute of Science, Bangalore}

\begin{document}
\maketitle
\thispagestyle{fancy}

\begin{abstract}
Human language has been described as a system that makes \textit{use of finite means to express an unlimited array of thoughts}. Of particular interest is the aspect of compositionality, whereby, the meaning of a compound language expression can be deduced from the meaning of its constituent parts. If artificial agents can develop compositional communication protocols akin to human language, they can be made to seamlessly generalize to unseen combinations. Studies have recognized the role of curiosity in enabling linguistic development in children. In this paper, we seek to use this intrinsic feedback in inducing a systematic and unambiguous protolanguage. We demonstrate how compositionality can enable agents to not only interact with unseen objects but also transfer skills from one task to another in a zero-shot setting: \textit{Can an agent, trained to `pull' and `push twice', `pull twice'?}.
\end{abstract}

\section{Introduction}
\label{section:introduction}
In the recent past, there has been a great deal of research in the field of emergent language in artificial agents interacting in simulated environments \cite{918430,HavrylovEtAl:2017:EmergenceOfLanguageWithMultiAgentGamesLearningToCommunicateWithSequencesOfSymbols,DBLP:journals/corr/abs-1912-06208,Gupta2020NetworkedMR}. However, the real question here is, to what extent do these evolved protocols resemble natural language? Recent studies have revealed the following about emergent languages: \textbf{(i)} they do not conform to Zipf’s Law of Abbreviation \cite{DBLP:journals/corr/abs-1905-12561}; \textbf{(ii)} communication protocols either do not follow compositionality patterns of natural language \cite{kottur-etal-2017-natural} or are not always interpretable~\cite{Lowe2019OnTP}; \textbf{(iii)} emerged protocols are sensitive to experimental conditions \cite{lazaridou2018emergence}.

Although compositionality is not crucial to achieving generalization, more compositional protocols have been shown to display higher zero-shot performance \cite{Ren2020Compositional}. While work on incorporating compositionality into emergent languages is still in its early stages, certain works \cite{DBLP:conf/aaai/MordatchA18,chaabouni2020compositionality} have proposed to use limited channel capacity as a means to achieve composition. However, we argue that agents may fail to develop meaningful communication protocols in such a restricted setting. Motivated by human behavior, we formulate intrinsic rewards~\cite{gopnikIM,baldassarreIM} to provide incentives to the agents for paying attention to communication despite having a limited channel capacity. Forced to deal with it to earn more intrinsic rewards, the agents must learn to use a more systematic and unambiguous protolanguage.

As proof of concept, we push the boundaries of compositionality to a more challenging multi-task settings, arguing that it can also support the acquisition of a more complex repertoire of skills (performing a \textit{pull twice} task when it has been trained to \textit{pull, push and push twice}), in addition to generalizing over novel composition of object properties (pushing \textit{red square} when it has been trained to push a \textit{red circle} and a \textit{blue square})
\footnote{\textbf{demos}:\href{https://sites.google.com/view/compositional-comm}{https://sites.google.com/view/compositional-comm}}. 


\section{Problem Setup}
\label{subsection: problem setup with emergent communication}
We analyze a typical signalling game \cite{Lewis1969-LEWCAP-4}, comprising a stationary Speaker-Bot (\textit{speaker}) and a mobile Listener-Bot (\textit{listener}), by modelling it in form of a Markov Decision Process specified by the tuple $(\mathcal{S}, \mathcal{O}, \mathcal{A}, r, \mathcal{T}, \gamma)$. At the beginning of each round, the speaker receives a natural language instruction (\textit{push a red circle}) and communicates the same using discrete messages $m_{i=1}^{n_m}$, sampled from a message space $\mathcal{M}$, to the listener over a communication channel. Here, $d_m$ is the dimension of the message $m_i$, and $n_m$ is the number of messages (these constitute the channel capacity, $|\mathrm{C}| = \mathrm{c}_{d_m}^{n_m}$). At each step $t$, the listener receives an observation $\mathbf{o}^{(t)} \in \mathcal{O}$, comprising the 2D grid-view and the received messages $m_{i=1}^{n_m}$, and takes an action $\mathbf{a}^{(t)} \in \mathcal{A}$. The goal of the listener is to choose optimal actions according to a policy $\bm{\pi} : (\mathcal{O},m_{i=1}^{n_m}) \mapsto \Delta(\mathcal{A})$, to maximize its long-term reward $\mathcal{R} = \sum_{t} \gamma^t r^{(t)}$. Here, $\gamma$ is the discount factor and $\mathcal{T}$ is the transition function $\mathcal{T} :\mathcal{S} \times \mathcal{A} \mapsto \mathcal{S}$. The environment generates a 0-1 (sparse) reward, i.e., the listener gets a reward of $r = 1$ if it achieves the specified task, otherwise $r = 0$. However, the listener has no information about either the task or the target, and relies on the speaker for the same. Given a language $\mathcal{L}(.) : \mathcal{C} \mapsto \mathcal{M}$, we use topographic similarity (\textit{topsim}) \cite{10.1162/106454606776073323} between $\mathcal{C}$ (set of concepts) and $\mathcal{M}$ (set of messages) as a measure of compositionality. Our work is contrasted with that of gSCAN \cite{ruis2020benchmark} which focuses on rule-based generalization using a supervised learning framework.

\label{subsection:compositionality}

\section{Approach}
\label{section:approach}

\subsection{Environment Description}
\label{subsection:environment description}



In our experiments, we use a $4 \times 4$ grid. Cells in the grid contain objects characterized by certain attributes like shape, size, color and weight. These objects can either be the \textit{target} object or the \textit{distractor} objects. Distractors have either the same color or the same shape (or both) as that of the target. We keep the number of distractors fixed ($=2$). The listener and the objects may spawn at any random location on the grid. Given an instruction, it is first processed using a parser to $\langle\mathrm{VERB}, \{\mathrm{ADJ}_i\}_{i=1}^{3}, \mathrm{NOUN}\rangle$\footnote{$\mathrm{VERB}$: task (`walk', `push', `pull'); $\mathrm{ADJ}$: object attributes like color (`red', `blue', `yellow', `green'), size (`small', `big') and weight (`light', `heavy'); $\mathrm{NOUN}$: object shape (`square', `circle', `cylinder', `diamond')}. The speaker transmits the same using a set of messages to the listener which, then, processes the grid representation and the received messages to achieve the given task. In our experiments, we use a $\{0,1\}^{d_{grid}\times 4\times 4}$ vector array for the grid representation, where each cell has a $d_{grid}$-dimensional encoding.

\begin{figure}[t]
	\centering
		\includegraphics[width=\linewidth]{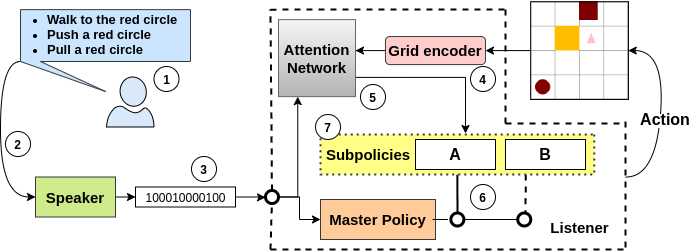}
		\caption{\small Model Description}
    \label{figure:model}
\vspace{-0.5cm}
\end{figure}

\begin{figure*}
	\centering
		\includegraphics[width=0.9\linewidth, height=3.7cm]{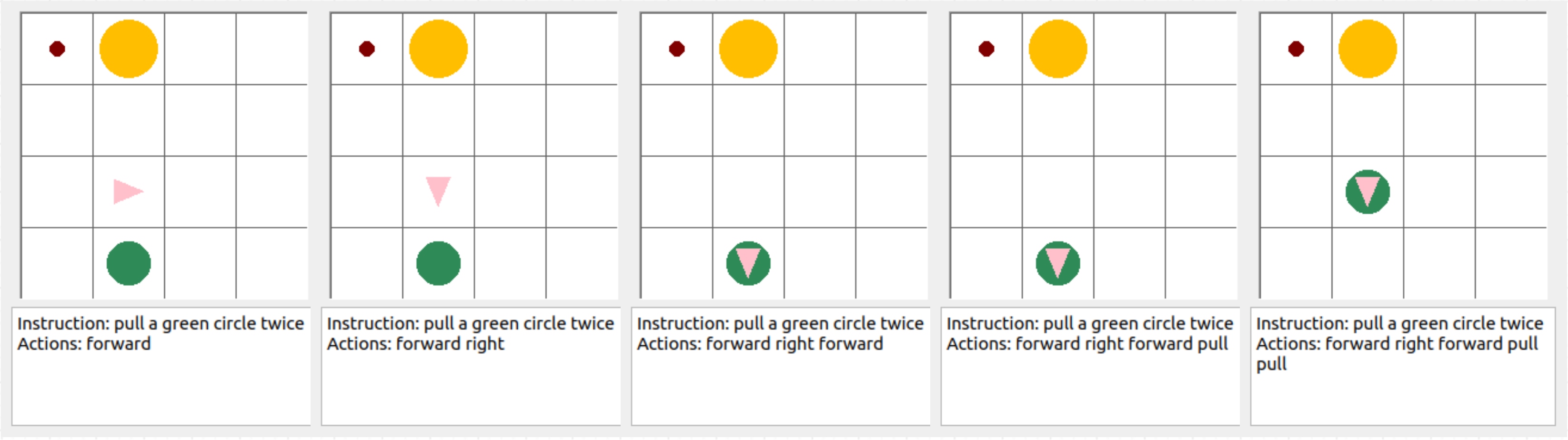}
		\caption{\small \textbf{[Best viewed in color]} Demonstration of Intrinsic Speaker on the numeral split for task PULL TWICE. Here, the green circle is heavy, hence the listener has to apply two units of force (TWICE) to pull it.}
    \label{figure:push_twice}
\vspace{-0.5cm}
\end{figure*}

\subsection{Model Description}
\label{subsection: model description}
(1,2) The speaker receives the parsed input instruction parsed. (3) The speaker uses an encoder to map the concept input to a hidden representation $\in \mathbb{R}^{n_m \times d_h}$. From this representation, a set of one-hot encoded messages $m_{i=1}^{n_m} \in \{0, 1\}^{d_m}$ are sampled (during training) using Categorical sampling, which are then transmitted over the communication channel. The number of messages $n_m$ is set to $|\langle\mathrm{VERB}, \{\mathrm{ADJ}_i\}_{i=1}^{3}, \mathrm{NOUN}\rangle|$. During evaluation, sampling is replaced with an $\argmax(.)$. We use the Straight Through trick \cite{JangEtAl:2017:CategoricalReparameterizationWithGumbelSoftmax} to retain differentiability. (4) At each step, the grid input is mapped in the Grid Encoder to an output $\mathcal{G}_t$ $\in \mathbb{R}^{d_{\mathcal{G}} \times 4 \times 4}$. (5) Next, we compute the attention weights $\alpha_{i=1}^{16}$ for each grid cell by taking a normalized dot product between $z$ and $\mathcal{G}_t^{i \times d_{\mathcal{G}}}$. A weighted combination is fed to the sub-policy networks. 

The listener learns to \textbf{(i)} identify the target object in the context of distractors, \textbf{(ii)} interact with the target object by executing a task specified by the speaker. We use a hierarchical-RL framework \cite{SUTTON1999181} for training. There are two sub-policies corresponding to the PUSH and the PULL tasks. (6, 7) In each round, the master policy selects either sub-policies using the received messages\footnote{actions spaces: master policy: \{A, B, Null\}; subpolicy A/B: \{left, right, forward, backward, push/pull\}}. The sub-policies have a shared input, which includes the grid encoder and the attention network. The whole framework is trained end-to-end using REINFORCE \cite{Williams:1992:SimpleStatisticalGradientFollowingAlgorithmsForConnectionistReinforcementLearning}.

In order to induce a more efficient training, we keep a measure of the Learning Progress (LP) of the listener for all tasks on a \textit{held-out} set, where LP for task $i$ is given as $\mathrm{LP}_i = | r_i - \mu_i |$. Here, $\mu_i$ denotes the running mean of rewards for task $i$. The tasks are sampled from a Categorical distribution with probabilities $p(i) = \frac{\mathrm{LP}_i}{\sum_j \mathrm{LP}_j}$ and, consequently, episodes corresponding to the sampled tasks are generated. This way, the listener can keep track of goals that are already learned, or can insist on goals that are currently too hard. 

\subsection{Inducing Compositionality}
\label{subsection:inducing_compositionality}
We would ideally want the concept to message mapping to be \textit{injective} (one-to-one), i.e. $\forall c, \tilde{c} \in \mathcal{C}, \mathcal{M}(c) = \mathcal{M}(\tilde{c}) \implies c = \tilde{c}$. Furthermore, the messages in $\mathcal{M}$ must exhibit a systematic structure (in holistic languages, one can satisfy the injective property without being compositional). Studies on language evolution have proposed limiting the channel capacity of communication as a constraint for achieving compositionality \cite{Nowak8028}. Yet, in the course of our experiments, on increasing  $|\mathcal{C}|$, we observed rather predictably that, with a limited channel capacity, it becomes increasingly difficult for the speaker to converge upon a consistent and unambiguous mapping from $\mathcal{C}$ to $\mathcal{M}$. Consequently, the listener would either ignore the information from the speaker (\textit{speaker abandoning}), or may exploit the inadequate information (\textit{undercoverage}\footnote{Inspired by machine translation works \cite{tu-etal-2016-modeling}, we define coverage as a mapping from a particular concept element to its appropriate message element. Full coverage refers to a distinct mapping of the whole concept input to corresponding symbols in $\mathcal{M}$.}) to converge on a local optimum (learning a fixed sequence of actions). To that end, we propose two types of intrinsic rewards to address these issues.

\paragraph{Undercoverage:} The limited channel capacity acts as an information bottleneck, impeding the speaker's ability to transmit  unambiguously. Therefore, it becomes difficult for the listener to infer the decoded messages at its end. To address this issue, we formulate a notion of compositionality from recent works in disentanglement \cite{Higgins2017betaVAELB}. We propose to use the Mutual Information (MI) between the concepts and the messages $\mathrm{I}(\mathcal{C}, \mathcal{M})$ as an intrinsic reward:

\vspace{-0.3cm}
\begin{align*}
\begin{split}
  \mathrm{I}(\mathcal{C}, \mathcal{M}) &= \mathrm{H}(\mathcal{C}) - \mathrm{H}(\mathcal{C} | \mathcal{M}) \\
    &= \mathrm{H}(\mathcal{C}) + \mathds{E}_{c \sim \mathcal{C}, m \sim \mathcal{M}(c)} \log p(c | m)
\end{split}
\end{align*}

Given that the training episodes are generated independent of the object specifications, $\mathrm{H}(\mathcal{C})$ can be assumed to be constant. We approximate the last term using Jensen's inequality \big($\mathds{E}_{c \sim \mathcal{C}, m \sim \mathcal{M}(c)} \big[\log p(c | m)\big] \geq \mathds{E}_{c \sim \mathcal{C}, m \sim \mathcal{M}(c)} \big[\log q_{\phi}(c | m)\big]$\big) to obtain a lower bound for $\mathrm{I}(\mathcal{C}, \mathcal{M})$.
Here, $q_{\phi}(c | m)$ is a learned discriminator module which takes the (concatenated) messages and tries to predict the concept labels (i.e. elements of $\langle\mathrm{VERB}, \{\mathrm{ADJ}_i\}_{i=1}^{3}, \mathrm{NOUN}\rangle$) and $\mathds{E}_{c \sim \mathcal{C}, m \sim \mathcal{M}(c)} \log q_{\phi}(c | m)$ is its negative cross-entropy loss. The final intrinsic reward is:

\vspace{-0.3cm}
\begin{equation}
    \mathrm{I}(\mathcal{C}, \mathcal{M}) \geq \mathrm{H}(\mathcal{C}) + \mathds{E}_{c \sim \mathcal{C}, m \sim \mathcal{M}(c)} \log q_{\phi}(c | m)
\label{equation:intrinsic_1}
\end{equation}

Intuitively, it suggests that it should be easy to infer the concepts from the messages. Conversely, the confusion (high error) arising from the speaker's inability to express concepts will lead to lower rewards. Note, that the reward will be highest when the conditions of full coverage and one-to-one mapping are satisfied (the discriminator will then be able to predict all the concept elements with high probability). We add the $\mathrm{I}(\mathcal{C}, \mathcal{M})$ reward at the last step of the episode, given as: $r[-1] + \lambda_1 \mathrm{I}(\mathcal{C}, \mathcal{M})$, where $\lambda_1$ is a tunable hyperparameter. The discriminator $q_{\phi}$ is periodically trained using batches sampled from a memory buffer, where we store the pair $\langle c_i,m_i \rangle$. Note, that we block the discriminator gradients to the speaker and use it merely as an auxiliary means to provide intrinsic feedback. 

\paragraph{Speaker Abandoning}
Existing works~\cite{Lowe2019OnTP} have shown that while training RL-agents augmented with a communication channel, it is likely that the speaker fails to influence the listener's actions. 
To address this, we propose to add another intrinsic reward to maximize the mutual information between the speaker's messages and the listener's actions, given the grid information.

At each step, we simulate $k$ intermediate steps to sample pseudo messages $\tilde{m}$ from the message distribution $\mathcal{M}$. Together with the original message $m$, we compute two sets of probability values corresponding to actions of the listener: (i) $\bm{\pi}(a_t | m, \mathcal{G}_t)$ which is the listener's policy conditioned on both the messages and the output of the grid encoder $\mathcal{G}_t$; (ii) $p(a_t | \mathcal{G}_t)$ or the probability distribution over the listener's actions conditioned on just the output of the grid encoder. We then calculate the mutual information for each step as follows:

\vspace{-0.35cm}
\small
\begin{align*}
    \mathrm{I}(a_t, m | \mathcal{G}_t) &= \sum_{a_t, m} p(a_t, m | \mathcal{G}_t) \log \frac{p(a_t, m | \mathcal{G}_t)}{p(a_t | \mathcal{G}_t)p(m | \mathcal{G}_t)} \\
    &= \sum_{a_t, m} p(m | \mathcal{G}_t) p(a_t | m, \mathcal{G}_t) \log \frac{p(a_t | m, \mathcal{G}_t)}{p(a_t | \mathcal{G}_t)} \\
    &= \mathds{E}_{m \sim \mathcal{M}} [\mathrm{D}_{KL}(p(a_t | m, \mathcal{G}_t)|| p(a_t | \mathcal{G}_t))]
\end{align*}
\normalsize

Note that $p(m | \mathcal{G}_t) = p(m)$ since messages and grid-view are independently processed. Here $p(a_t | \mathcal{G}_t)$ is obtained by marginalizing over the joint probability distribution, given as, $\sum_{\tilde{m}} p(a_t, \tilde{m} | \mathcal{G}_t) = \sum_{\tilde{m}} p(a_t | \tilde{m}, \mathcal{G}_t) p(m)$. We use Monte Carlo approximation to replace the Expectation by sampling messages from $\mathcal{M}$. The final reward equation for $k$ pseudo-steps is given as:

\vspace{-0.35cm}
\small
\begin{multline}
\label{equation:intrinsic_2}
    \mathrm{I}(a_t, m | \mathcal{G}_t) \\
= \frac{1}{k} \sum_{m} \mathrm{D}_{KL} \big[\bm{\pi}(a_t | m, \mathcal{G}_t) || \sum_{\tilde{m}} \bm{\pi}(a_t | \tilde{m}, \mathcal{G}_t) p(m)\big]
\end{multline}
\normalsize

Maximizing Equation~\ref{equation:intrinsic_2} leads to a higher speaker influence on the listener's actions. The net reward at each step is given as: $r_t + \lambda_3 \mathrm{I}(a_t, m | \mathcal{G}_t)$, where $\lambda_3$ is a tunable hyperparameter. 

\section{Experiments}
\label{section:experiments}

\paragraph{Zero-Shot Generalization Splits:}
\textbf{(i)} \textbf{Visual split}: All episodes not containing the `red square' as a target object, were used for training the model. During evaluation, we examine whether the trained model can generalize to the following instructions: \textit{walk to a red square}; \textit{push/pull a red square}. \textbf{(ii)} \textbf{Numeral split}: The training set contains instructions with \textit{Push}, \textit{Push Twice} and \textit{Pull}, whereas, test set contains \textit{Pull Twice} task. Here the modifier \textit{Twice} is used to denote a heavier object i.e., listener should execute two consecutive `pull' actions to move the object. The listener must infer from its training that a symbol corresponding to \textit{heavy} requires twice as many actions.


\begin{figure}[t]
\centering
    \includegraphics[width=\linewidth]{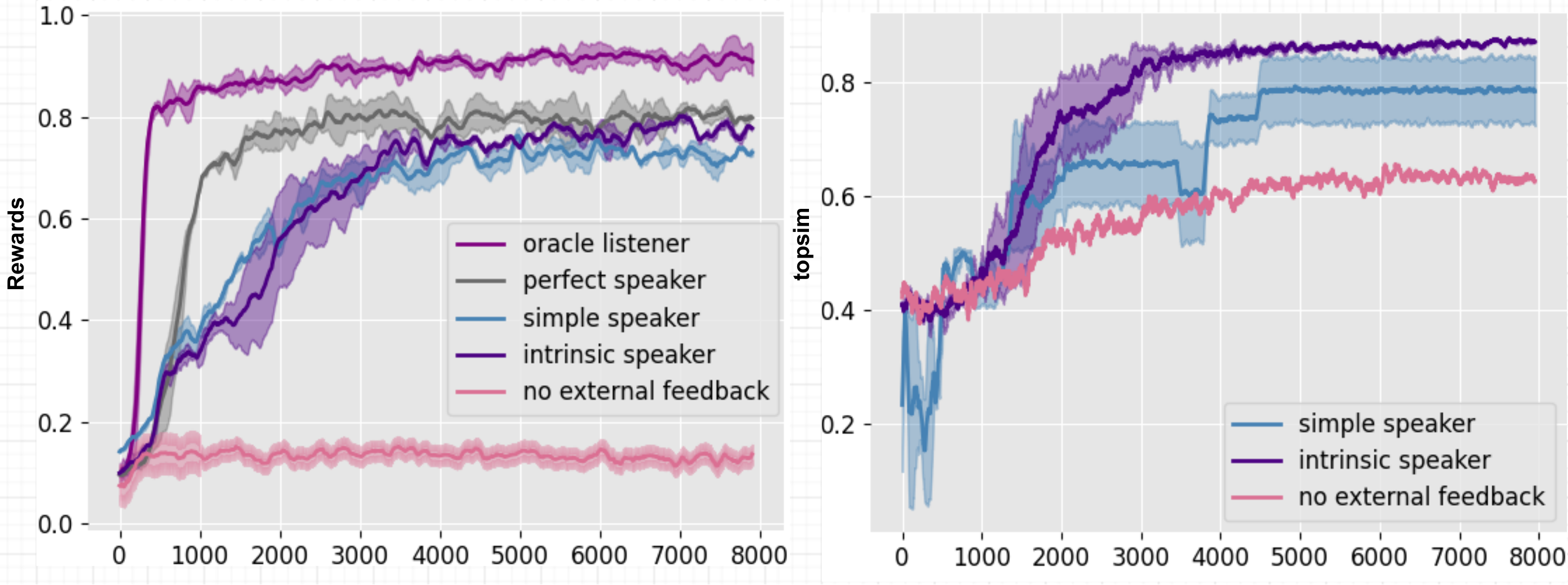}
		\caption{\small \textbf{[Best viewed in color]} \textbf{Left}: Comparison of Intrinsic Speaker with other baselines on a single policy module for WALK task. It can be observed that Intrinsic Speaker performs as well as the Perfect Speaker baseline; \textbf{Right}: Comparison of topsim metric of Intrinsic Speaker (with and without feedback) and Simple Speaker. All plots have been obtained by averaging the validation rewards obtained over 5 independent runs. [X-axis: 1 unit = 50 episodes]}
    \label{figure:walk_comparison}
\vspace{-0.40cm}
\end{figure}

\begin{table}[t]
	\small
	 \begin{center}
		 \begin{tabular}{ >{\centering\arraybackslash}m{2.9cm} >{\centering\arraybackslash}m{2.2cm} >{\centering\arraybackslash}m{1.7cm}}
			 \toprule
			 \textbf{Task} & \textbf{Model} & \textbf{Zero-Shot Accuracy}\\[0.4ex] 
			 \midrule
			 \textit{walk to a red square} (visual split) & \begin{tabular}{>{\centering\arraybackslash}m{2.2cm}>{\centering\arraybackslash}m{1.5cm}} Simple Speaker & $73.43 \%$ \\ \midrule Intrinsic Speaker & $\mathbf{80.24 \%}$\end{tabular}\\
			 \midrule
			 \textit{push a red square} (visual split) & \begin{tabular}{>{\centering\arraybackslash}m{2.2cm}>{\centering\arraybackslash}m{1.5cm}} Simple Speaker & $67.17\%$ \\ \midrule Intrinsic Speaker & $\mathbf{72.45}\%$\end{tabular}\\
			 \midrule
			 \textit{pull a red square} (visual split) & \begin{tabular}{>{\centering\arraybackslash}m{2.2cm}>{\centering\arraybackslash}m{1.5cm}} Simple Speaker & $66.80\%$ \\ \midrule Intrinsic Speaker & $\mathbf{73.29}\%$\end{tabular}\\
			 \midrule
			 \textit{pull a red square twice} (numeral split) & \begin{tabular}{>{\centering\arraybackslash}m{2.2cm}>{\centering\arraybackslash}m{1.5cm}} Simple Speaker & $65.25\%$ \\ \midrule Intrinsic Speaker & $\mathbf{69.77}\%$\end{tabular}\\
			 \bottomrule
		 \end{tabular}
	 \end{center}
	 \caption{\small Comparison of simple speaker and intrinsic speaker zero-shot performance on different splits.}
	 \label{tab_results1}
\vspace{-0.5cm}
 \end{table}
 
\paragraph{Baselines:} We compare our Intrinsic Speaker model with the following baselines. \textbf{(i) Oracle Listener}: For each cell, we zero-pad the grid encoding with an extra bit, and set it ($=1$) for the cell containing the target object. This way, the listener has complete information about the target in context of the distractors. We use this baseline as our upper limit of performance. \textbf{(ii) Perfect Speaker:} The speaker uses an Identity matrix that channels the input directly to the listener. Thus, it is perfectly compositional. \textbf{(iii) Simple Speaker:} Here the speaker-listener is trained end-to-end without using the intrinsic rewards. 

\section{Results}
\label{section:results}

\textbf{(i)} The proposed Intrinsic Speaker outperforms the Simple Speaker in terms of both, convergence rewards and topsim score (Figure~\ref{figure:walk_comparison}). In fact, the Intrinsic Speaker matches the performance of the Perfect Speaker, thus, showing that the emergent communication is highly compositional ($\approx 0.9$).  \textbf{(ii)} The zero-shot generalization accuracy in Table~\ref{tab_results1} shows that the Intrinsic Speaker consistently outperforms the Simple Speaker on both splits. \textbf{(iii)} In order to test the effectiveness of intrinsic rewards in inducing compositionality, we trained the Intrinsic Speaker with no external reward from the environment. As shown in Fig~\ref{figure:walk_comparison} (right), the intrinsic rewards were alone capable of generating a topsim score $\approx 0.6$. For further details, the readers are requested to refer the paper \cite{DBLP:journals/corr/abs-2012-05011}.


\bibliography{biblio}
\bibliographystyle{acl_natbib}
\end{document}